# Optimized Car Damage Detection using CNN and Object Detection Model

1st Sania Ejaz
School of Engineering & IT
Manipal Academy of Higher Education, Dubai Campus
Dubai, U.A.E
sania.ejaz@dxb.manipal.edu

2nd Jaison Shajan Thomas
School of Engineering & IT
Manipal Academy of Higher Education, Dubai Campus
Dubai, U.A.E
jaison.thomas@dxb.manipal.edu

3rd Zaheeruddin Ahmed
School of Engineering & IT
Manipal Academy of Higher Education, Dubai Campus
Dubai, U.A.E
zaheeruddin@manipaldubai.com

4th Sushma Hans
School of Engineering & IT
Manipal Academy of Higher Education, Dubai Campus
Dubai, U.A.E
sushma.hans@dxb.manipal.edu

***Abstract* — Car damage detection systems play a vital role in scenarios, like the insurance claims process, accident reporting systems, or a documented write-up of the various damages a vehicle in a car garage exhibits upon entry and exit. This paper presents a two-fold model architecture encompassing a convolutional neural network used to classify the collected input samples as clear or damaged, thereby filtering out the undamaged images, before the input is passed on to the object detection layer for further localization and classification of defects into Scratch, Dent, Glass Shatter, and Smash. A deviation from the focus of related work, this paper poses the use of unified models while applying transfer learning techniques to augment performance that caters to multiple use cases. The design of this model has been made to optimize performance by incorporating the two models, to streamline the processing of input samples. The model boasts significant performance with accuracies of 98% for the CNN layer and 0.5 mAP for the object detection layer. However, the metric for the object detection layer seems to understate its performance due to reasons discussed within the paper.**

## I. INTRODUCTION

Reportedly, around 13 car accidents occur every minute [1], resulting in an enormous surge in insurance claims. The staggering rate of car accidents imposes a distressing strain on insurance companies in the form of insurance claims leakage. Advances in accident reporting systems have both parties capture the damages as inputs to their reporting system, which is then forwarded to insurance companies to estimate damages and costs. Mall parking and other institutions now capture cars entering their premises to facilitate number plate recognition, thereby collecting data that can be channeled to propagate customer scouting for garages and companies on the identification of car damages, enabling targeted marketing campaigns. Second-hand car deals can benefit from an automatic remote vehicle inspection that safeguards the buyer's interests.

The above-mentioned use cases reinforce the yields of an efficient Car Damage Detection system that automates the cumbersome process of vehicle inspection, not only mitigating time and effort but also alleviating costs.

Gaining significant popularity over the decades, Deep Learning, a subset of Machine Learning and Artificial Intelligence, evinces a powerful ability to accommodate the increased complexity of the problem to be modeled, with its multitude of neural layers. This, coupled with its proficiency to consume large volumes of data, even unstructured, while applying autonomous feature extraction, is what makes it a go-to approach for modeling applications such as car damage detection.

Developing an application that tackles the need for an efficient system as such is touted to be challenging, given its nature of largely overlapping class labels pertaining to different damages, which makes it difficult to segregate these detected defects into exclusively independent labels. Furthermore, given the specificity of the domain, there appears to be a lack of sufficient datasets available for training such a complex neural network. Factors like speed, computational expense, and the trade-off between latency and accuracy are also to be considered when building the solution. Therefore, while the development of such an application can prove to be extremely impactful to its user base, the complexity of building such a network needs to be methodically approached.

## II. RELATED WORK

Work in this domain [2, 3, 4, 5, 7] has been primarily focused on insurance claim leakage and has prominently taken up a similar approach of using segmentation with two-stage models like RCNN, Faster RCNN, and other variations. A few works presented deviations from this approach to draw up comparisons between different techniques of building CNN models [6].

[5] utilizes an RCNN network to extract ROI, which is subsequently segmented and processed by CNN models, to tag each segment with a label that indicates the presence or absence of a scratch. This approach is computationally expensive and makes the model dense while increasing the latency of a detection.

[6] evinces a multilevel analysis, extending to assess the severity of the damage detected in the previous stage. Their approach was structured to compare the performance of a transfer learning approach as opposed to fine-tuning domain-specific pre-trained CNN models. [4, 7] document their journeys of moving from building models from scratch to fine-tuning and finally settling for the transfer learning approach, yielding maximum returns.

While [3] proposes an improved variation of the Mask RCNN model, [2] utilizes the existing model coupled with transfer learning over various fine-tuned parameters to realize accuracy sufficient for their use cases.

Other significant work we reviewed during our research pertained to papers that presented a comparative analysis of existing models in the deep learning models collection [8, 10], notable additions to the model zoo [9], and techniques that are proven to augment the performance of a deep learning model [11, 12, 13]. The key insights derived from this research have aided us in structuring our approach and customizing the design of the model to best suit our purpose.

## III. APPROACH

Our approach was directed by correlating the insights gained from research with the challenges outlined previously while molding it to best represent our scenario.

Our use case expands not only to insurance claims leakage but also automatic inspection in garages, customer scouting in parking spaces, and second-hand car sales, alike, we recognize that this kind of application would be faced with many instances of undamaged cars, bringing in the need for a preliminary filter, tackled by a robust CNN classification model. The CNN model classifies the captured frames as either damaged or not. In the case that a damaged image is encountered, the frame would be passed to an Object Detection model for localization and classification of the defect.

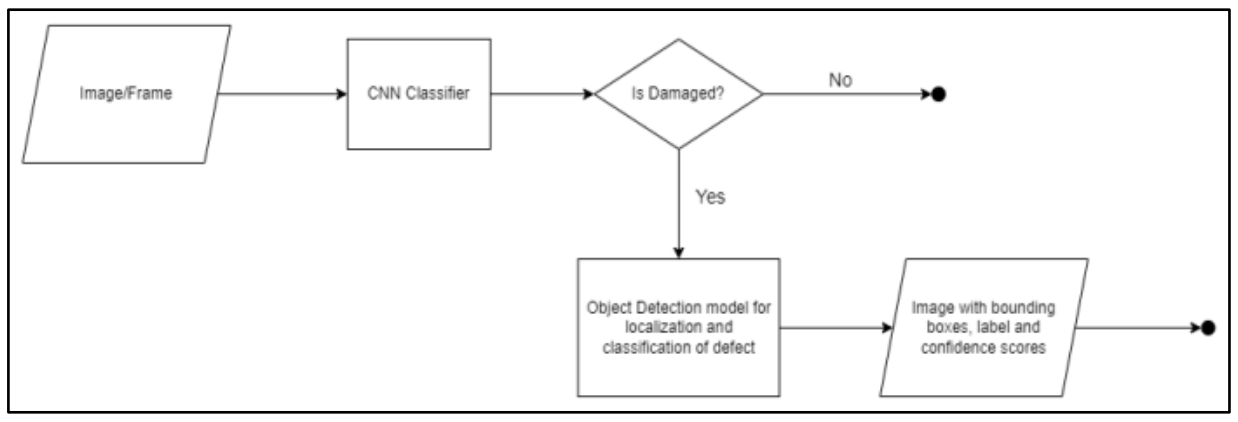

Fig. 1. Flowchart representation of our designed approach

As mentioned earlier, the application we are modeling is challenging and may have diverse labels to represent many kinds of defects, we are thereby limiting our scope to represent 4 categories, namely, glass shatters, scratches, dents, and smashes.

Next, to counter the issue of scarcity of sufficient datasets, to build optimized weights and a strong foundation for the network architecture, as well as mitigate the training time, we take to a very popular technique called transfer learning. As supported by [6], transfer learning eases the process of building dense networks and establishing a strong architecture by simply transferring the knowledge gained on operating on the source domain to facilitate similar tasks on the target domain [11]. This originates from the generalization theory of transfer, which states that "learning to transfer is the result of the generalization of the experience" [11].

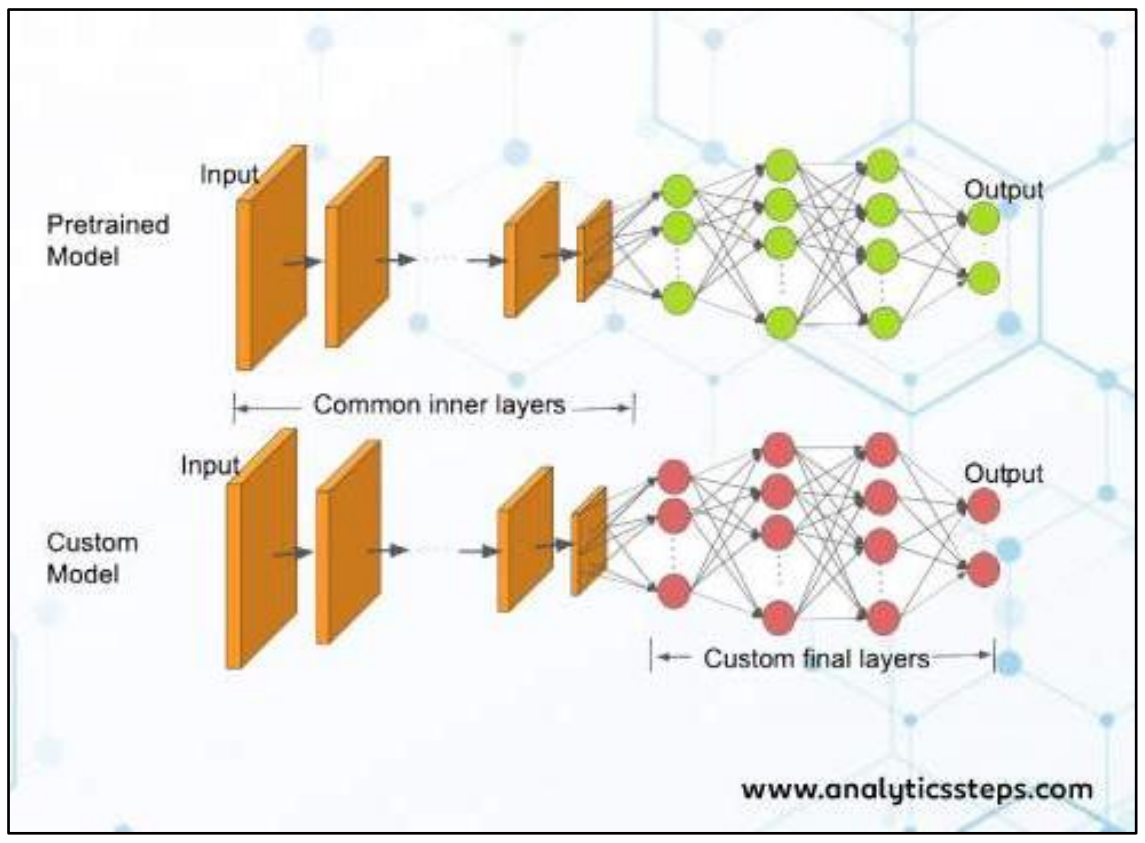

Fig. 2. Representation of the architectural designs of networks that implement transfer learning techniques. It is observed that the first few layers of such an architecture pose as the feature extractors, along with layers that overlap in both networks, that are retained and generalized to serve their purpose in the customized model. The final layers are tweaked to the specific task at hand with the target domain.

This ensures that the weights and architecture built on a different database with an enormous magnitude of samples can be downloaded and tweaked to perform a similar task on the customized dataset, that serves as the target domain, thereby assuring an improved performance with the limited dataset.

Furthermore, image augmentation techniques [12] can be applied throughout the dataset to generate variations in the sample to achieve a well-generalized model. With these decisions in place, we proceed to the next stage of data collection and model selection to address the other key factors listed in Section 1.

## IV. DATA COLLECTION

We sourced our dataset from Kaggle, collecting over 3000 images, containing images depicting the 4 categories of damages, as well as images representing the undamaged class. We then segregated these images into two folders, creating two datasets to train our 2 models separately.

Images of the undamaged class are exclusively shifted to the dataset pertaining to our CNN model that would act as the classifier to filter out the undamaged car captures passed as input. Creating a dataset for the damaged class of this model brings about an overlap in the images found in both the datasets of our CNN model as well as that of the Object Detection model.

Finally, we have around 1027 images set aside for training our Object Detection model and 200 images for the testing phase, representing scratches, smashes, glass shatters, and dents. On the other hand, our CNN dataset finds a similar distribution ratio of about 80:20 percent, resulting in around 2600 images in the train set and 640 in the test set, pertaining to both the classes of damaged and clear.

To prepare this dataset for our object detection model, we annotated its dataset using the labelImg software, which allows for efficient manual labeling of images.

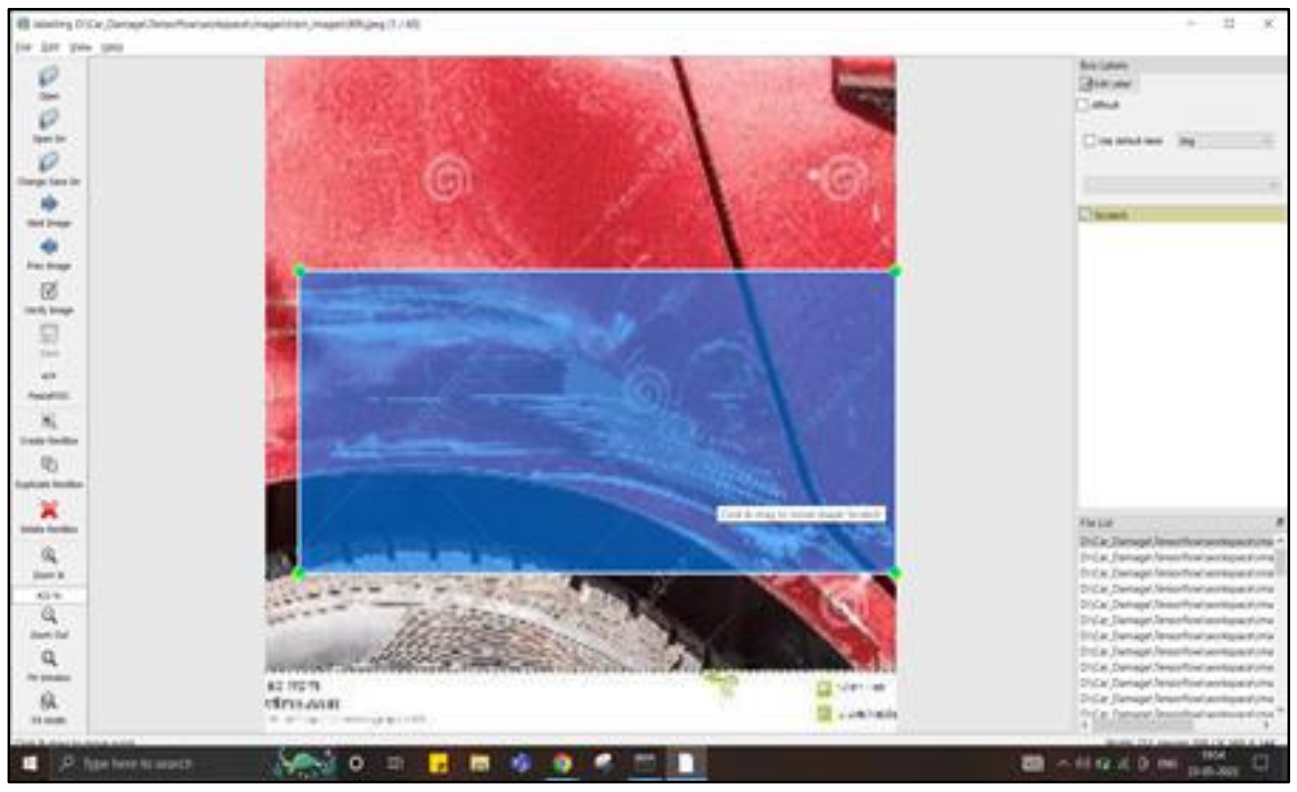

Fig. 3. Illustrates the process of annotating the dataset. In each image, bounding boxes are manually drawn around areas of damage to be captured, which are then linked to a corresponding label class. An image may contain multiple annotations if it includes several instances of damage. Each annotation is subsequently stored as an XML record

During the phase of collection and annotation, we made sure to include samples portraying diverse angles, lighting conditions, and colors, filtering the downloaded datasets to suit our purpose. Recognizing the impact of data augmentation on generalizing and enhancing the performance of a model such as ours, we applied different image transformation techniques to synthetically increase our training data.

The approach to applying these techniques differs between the two models. In the case of our CNN model, we simply load the dataset with Python's ImageDataGenerator method that performs an internal data augmentation, increasing the training size per epoch by rotating, flipping, and zooming into images. Whereas the pre-trained model requires us to modify its pipeline.config, containing all its configuration properties, passing predefined values to its data_augmentation_options, to which we have passed parameters to rotate, flip, crop, and adjust the brightness of images.

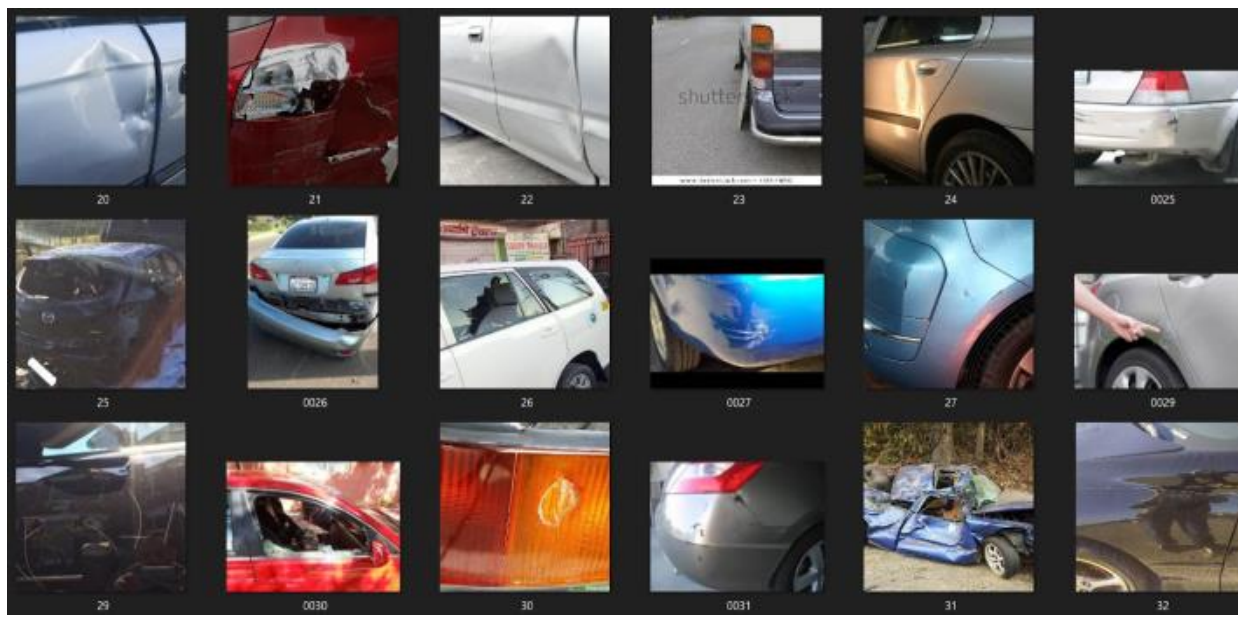

Fig. 3.1. Sample of collected data for our object detection model

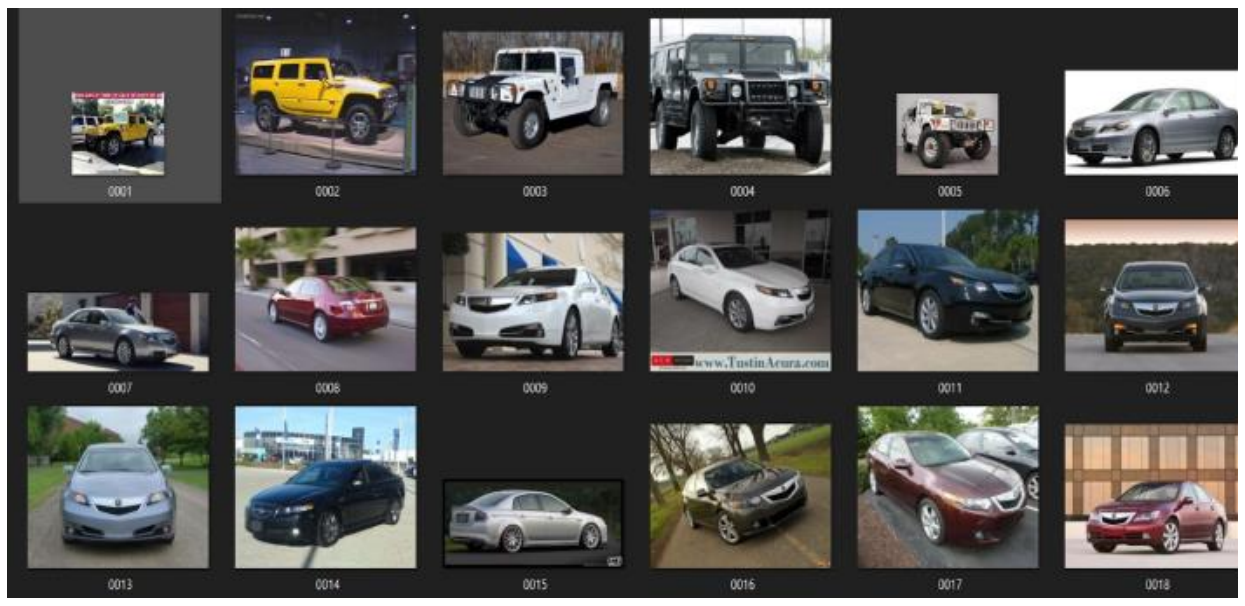

Fig. 3.2. Sample of collected data for our CNN classifier, representing out 'clear' class

## V. MODEL ARCHITECTURE

Our first objective is to build an image classifier with just two classes, one the negation of the other, which classifies this case into that of binary classification. Since our first objective is a simple classification model, we decided to build our own model.

### *A. Convolutional Neural Networks*

Convolutional Neural Networks (CNNs) are a prominent subtype of Neural Networks that are especially associated with processing unstructured data like images, videos, and audio. A CNN model has three fundamental layers to its core: The convolutional layer, the Pooling layer, and the Fully Connected layer.

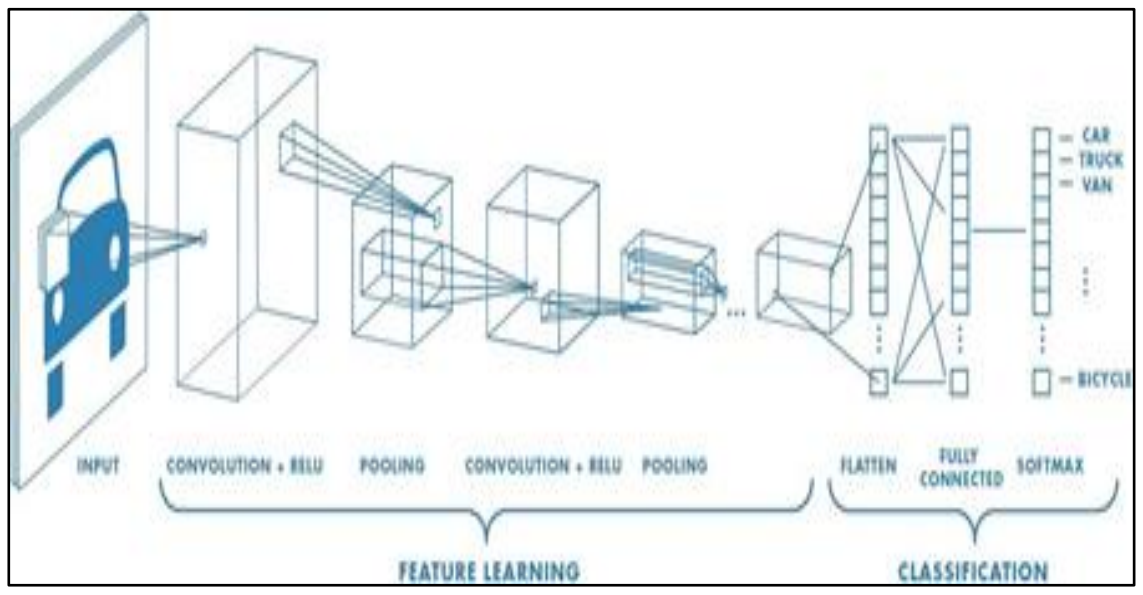

Fig. 4a. The model architecture of a Convolutional Neural Network

The convolution layer is at the heart of a CNN model and is composed of many kernels/filters that are used to extract features of the passed input image. Defined as a matrix, that moves over the input data in strides, a parameter that controls the amount of movement along the image or video, kernels perform the dot product with the sub-region of the input data thereby building a feature map that identifies features and defines their strength in their strides [15]. This map is passed to the Max pooling layer which optimizes the map to represent its most prominent features.

Proceeding to the Flatten layer, this feature map is converted to a linear array that would be fed to our Dense layer, tasked with building a fully connected network [16] that would perform the task of image classification.

### *B. Defect localization and classification: Coco SSD Mobnet*

To select the most appropriate pre-trained object detection model with the most suitable base network, we browsed various research papers and articles online to draw up our choices and finally decide upon the best suitable model. We took to the TensorFlow Model Zoo to choose our pre-trained model framework.

Deviating from the common approach of using two-staged models, we decided to utilize unified object detectors, owing to their speed and computationally cheaper nature, narrowing down our options to 2 popular models, namely YOLO and SSD. Considering the analysis performed in [8][10], SSD proves to be faster, and more accurate and evinces an ability to handle different object sizes, and accommodate end-to-end training.

Coming to our base network, we considered options like VGG16, ResNet, MobileNet, etc. Finally, we opted for MobileNet owing to the analysis performed in [9] that introduces a new class of efficient models for vision-based applications.

Single Shot Detector (SSD) comprises a deep neural network, that generates scores for every category of an object within each frame, proceeding to combine multiple prediction feature maps with diverging resolutions, which gives it its ability to work with different object sizes.

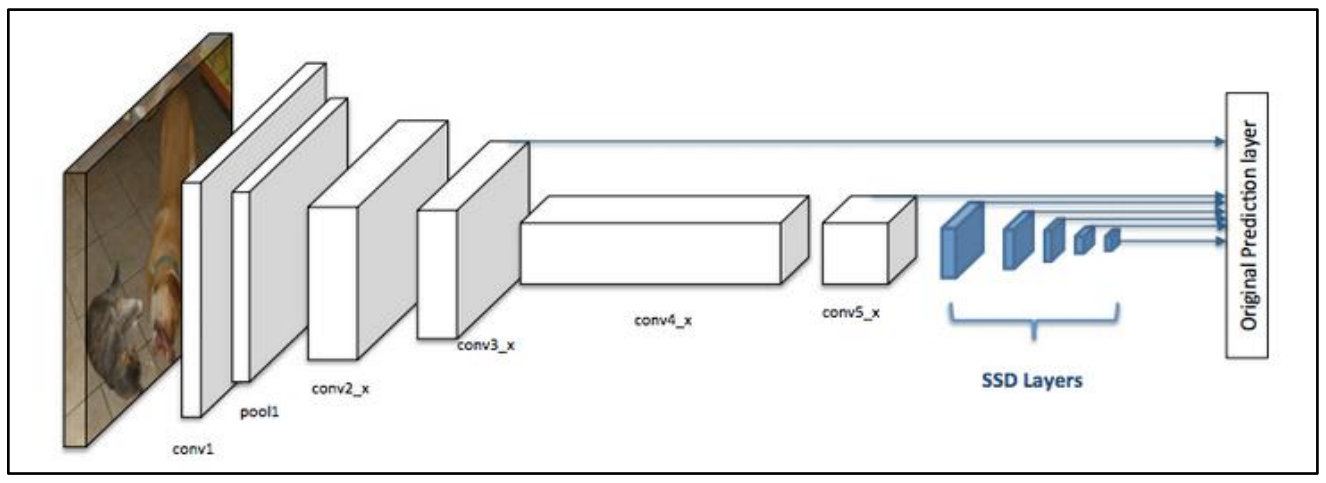

Fig. 4b. The model architecture of an SSD network

The pre-trained model is trained on the COCO Image dataset, which is a large-scale object detection, segmentation, and captioning dataset, comprising 80 object categories with 1.5 million object instances [17].

Finally, our base network, MobileNets, utilizes depthwise separable convolutions/factorized convolutions to build lightweight networks. These factorized convolutions split standard convolutions into 2 distinct layers, drastically reducing computation and model size [9].

The above mentioned layers are comprised of a depthwise convolution that applies a single filter on each input channel, while the following layer is a pointwise convolution, a 1*1 convolution, that combines the output of depthwise layers.

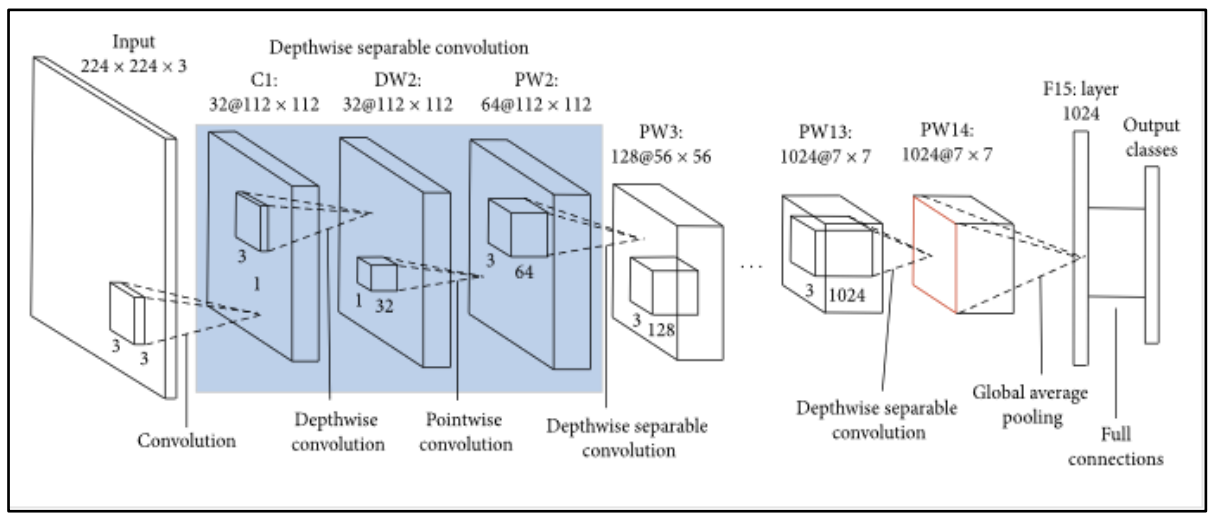

Fig. 4c. The model architecture of a MobileNet

## VI. TRAINING

### A. *Convolutional Neural Network (CNN) Classifier*

The model consists of a sequence of the core layers to a CNN model, followed by one hidden layer and an output layer. In the first layer, we define 32 filters of size 3*3, and a specified input size of 64*64. This is then followed by the Max Pooling layer, defined with a 2*2 pool size. Next, we add our Flatten layer.

The hidden layer is defined with 128 units to represent the output dimensionality, a random kernel initializer, and a Relu activation function. Finally, we add our output layer, with 2 units representing our classes of ‘clear’ and ‘damaged’, a random kernel initializer, with a Softmax activation function to normalize the probability distribution and circumvent the vanishing gradient descent issue.

| Attribute/Parameter | Value |
|---|---|
| **Tensorflow version** | 2.9.0 |
| **Keras version** | 2.9.0 |
| **Filters** | 32 |
| **Batch size** | 32 |
| **Steps per epoch** | 57 |
| **Validation steps** | 14 |
| **Pool size** | 2 |
| **Num classes** | 2 |
| **Epochs** | 90 |

Table 1: Model environment and parameter table for the CNN model

### B. *Object Detection Model*

While preparing the pre-trained object detection model to suit our purpose we had to replace its dataset, create a label map to associate the labels provided in annotations to the classes we define for our model, and then tune its configuration file before moving on to the training step. To consolidate our dataset, we converted them to TFRecords, which brings the model efficiency by reducing space complexity, allowing for parallel computation, and ensuring these records are self-contained.

In the pipeline.config, we customize the model configuration to meet our needs by altering the number of classes, resizing the input images, tweaking the data augmentation options, optimizers, and so on.

| Attribute/Parameter | Value |
|---|---|
| **Tensorflow version** | 2.8.0 |
| **Keras version** | 2.8.0 |
| **Num classes** | 4 |
| **Epochs** | 35000 |
| **Warmup steps** | 1000 |
| **Kernel size** | 3 |
| **Regularizer** | L2 |
| **Warmup learning rate** | 0.0266 |
| **Momentum optimizer** | 0.9 |

Table 2: Model environment and parameter table for the object detection model

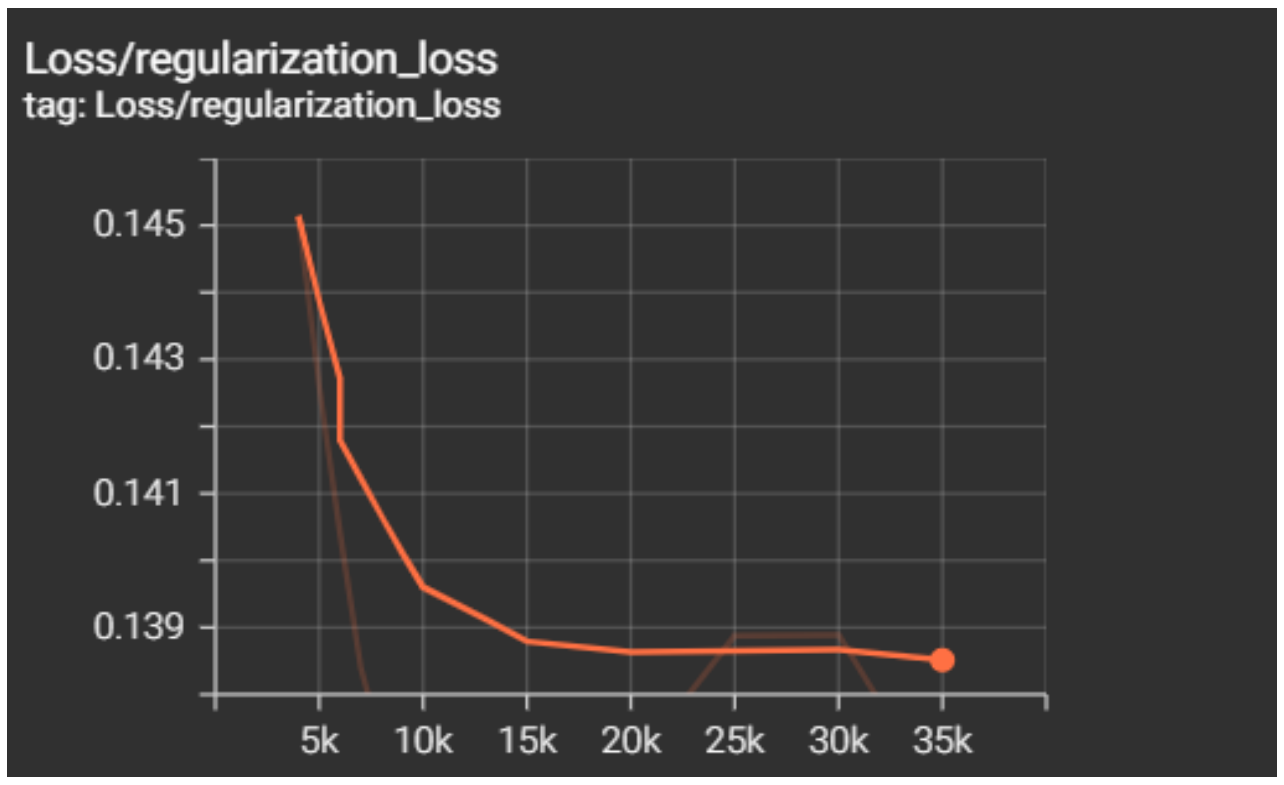

Fig. 5.1. Regularization loss curve captured through the tensorboard during training

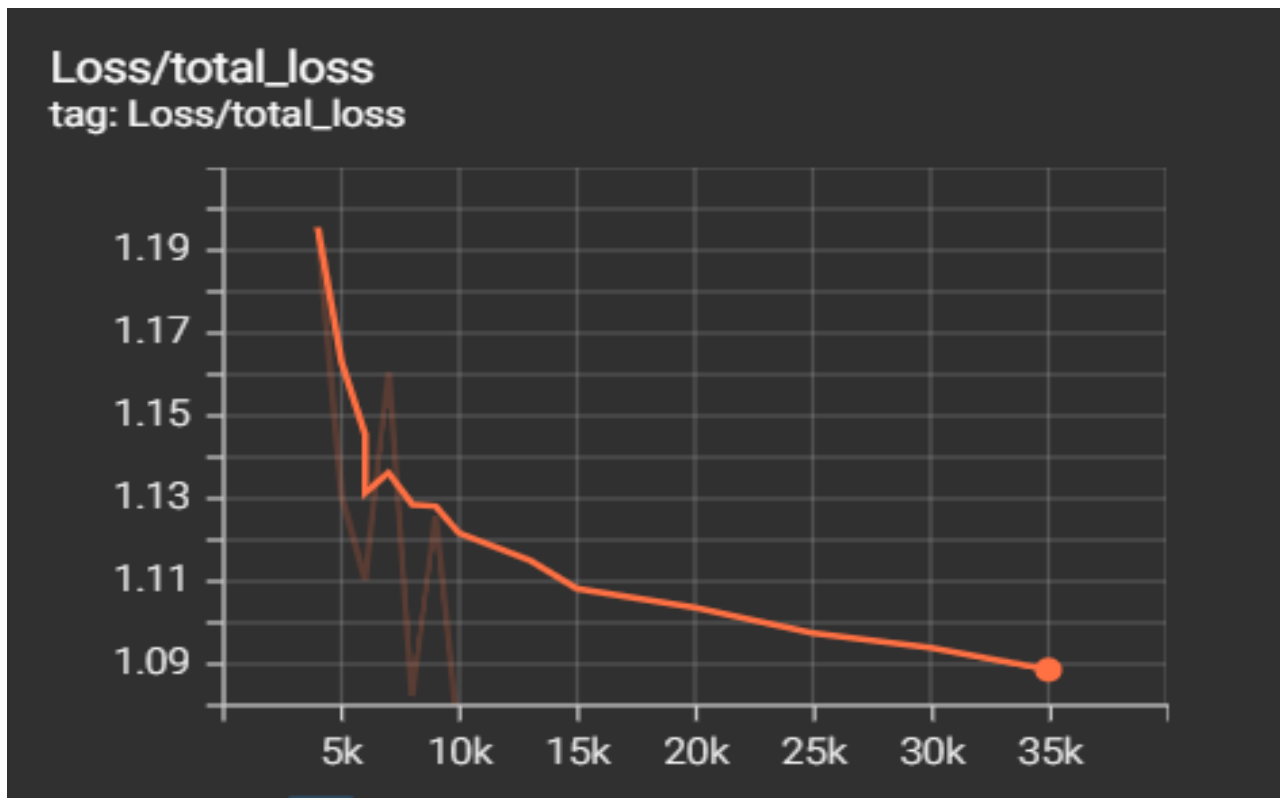

Fig. 5.2. Total loss curve captured through the tensorboard during training

## VII. RESULTS

This section displays the results achieved by our models, and analyzes their performance in context to its evaluation metrics, alongside the outputs presented by the models. The CNN classifier is evaluated against the accuracy metric, which measures the ratio of the total correct classification instances to that of the total number of test cases, and has amounted to a 98% accuracy.

The object detection model is assessed using two metrics, namely, precision which represents the fraction of relevant instances to that of retrieved, and recall which represents the fraction of relevant instances that were retrieved.

$$\text{Precision} = TP / (TP + FP)$$

$$\text{Recall} = TP / (TP + FN)$$

Where,
TP (True Positives): Model correctly predicts the positive class. [The model predicted the positive class, and the actual value was indeed positive.]

FP (False Positives): Model incorrectly predicts the positive class. [The model predicted the positive class, but the actual value was negative.]

FN (False Negative): Model incorrectly predicts the negative class. [The model predicted the negative class, but the actual value was positive.]

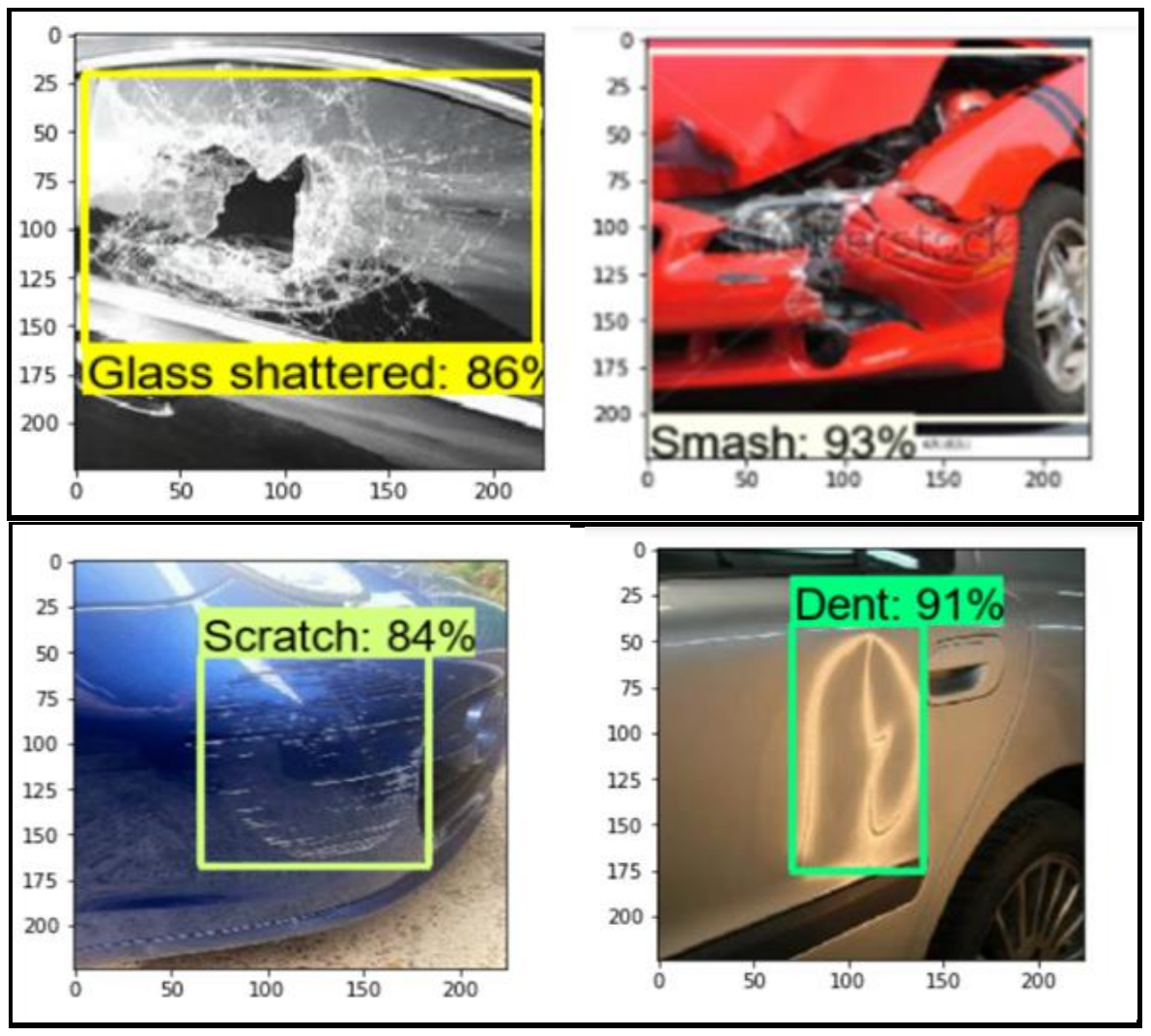

Fig. 6.1. The damages identified and localized by the object detection layer, starting from glass shattered, smash, scratch to dents in images with single instances

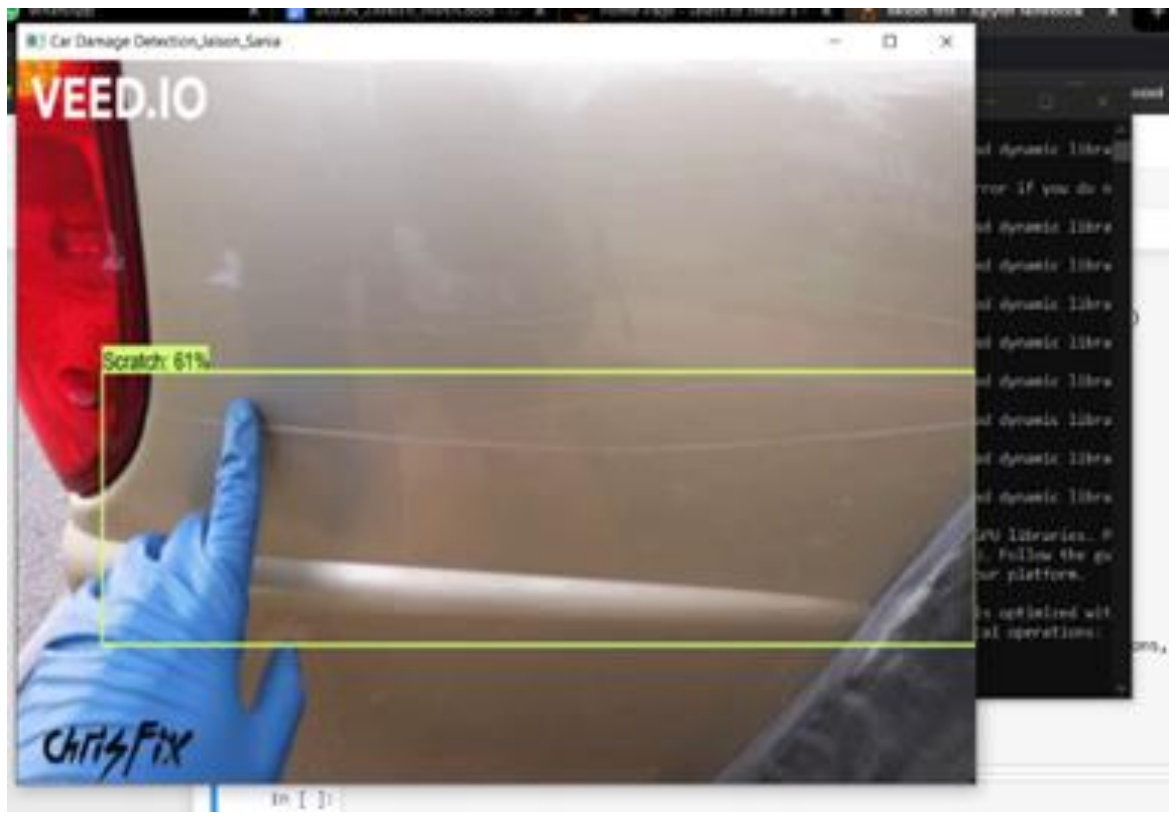

Fig. 6.2. The damages identified and localized by the object detection layer on a video feed, correctly identifying scratches.

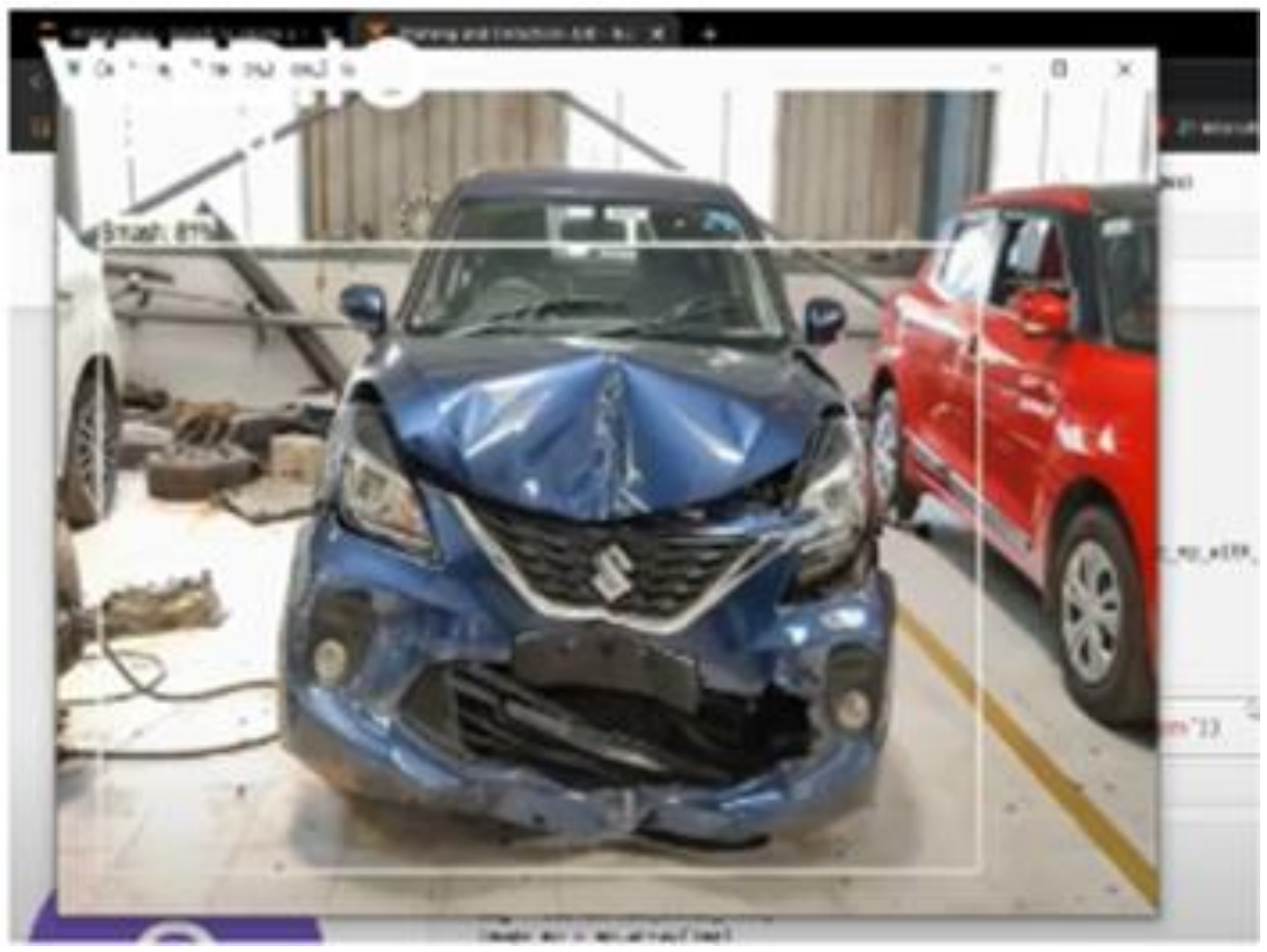

Fig. 6.3. The damages identified and localized by the object detection layer on a video feed, correctly identifying smashes.

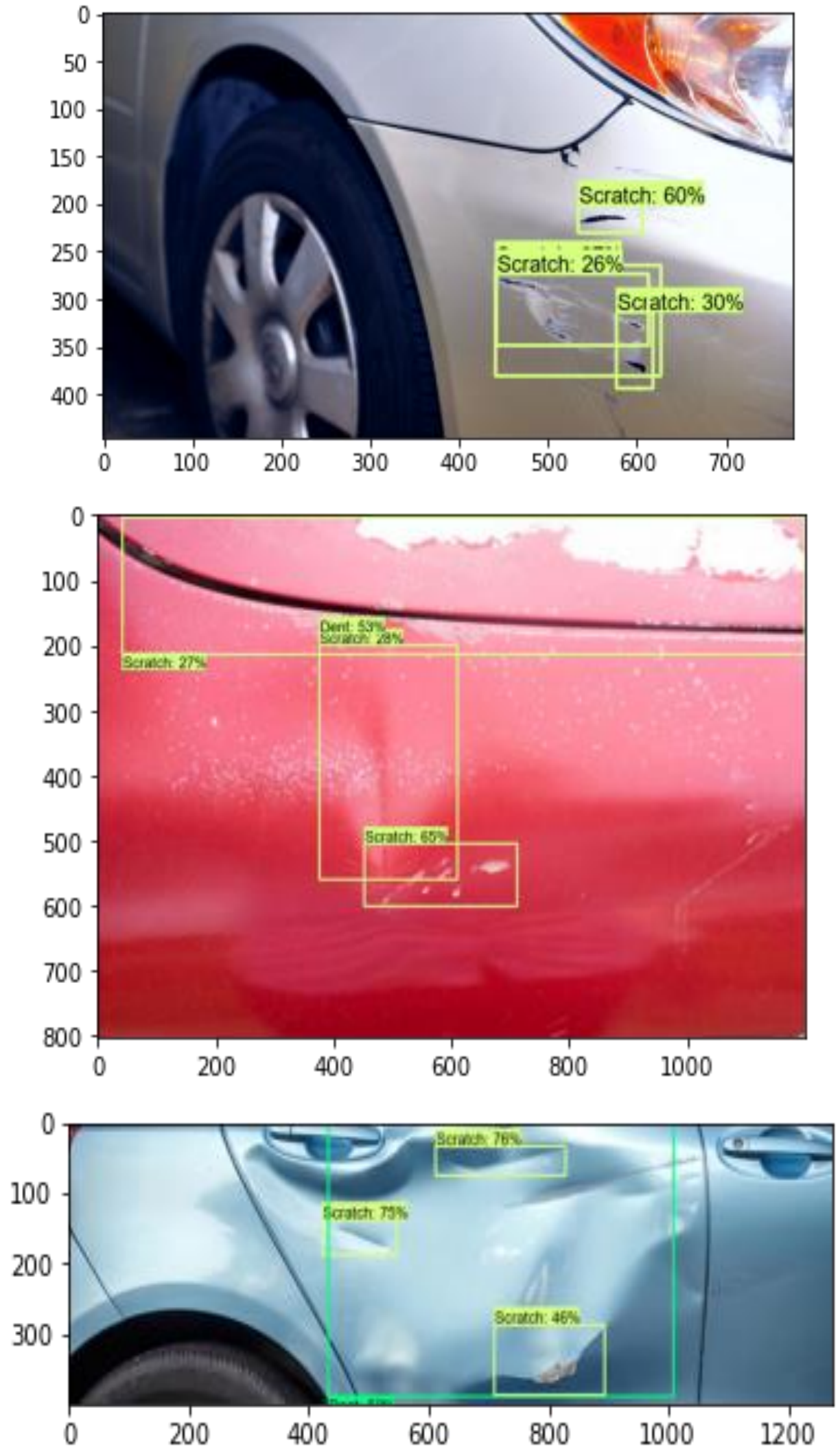

Fig. 6.4. The damages identified and localized by the object detection layer in images of multiclass instances. The model precisely locates multiple scratches in the first image, identifies a dent among scratches and in the last image, scratches among the dent.

To demonstrate the efficiency of our two-fold approach, we ran batches of test images with and without our CNN Classifier and captured the time stamps.

| Num of images | With CNN | Without CNN |
|---|---|---|
| **10** | 7.6628s (4 clear images) | 39.0248s |
| **30** | 14.6243s (16 clear images) | 139.4886s |

Table 3: Model results with and without classifier to demonstrate the efficiency of our two-fold approach

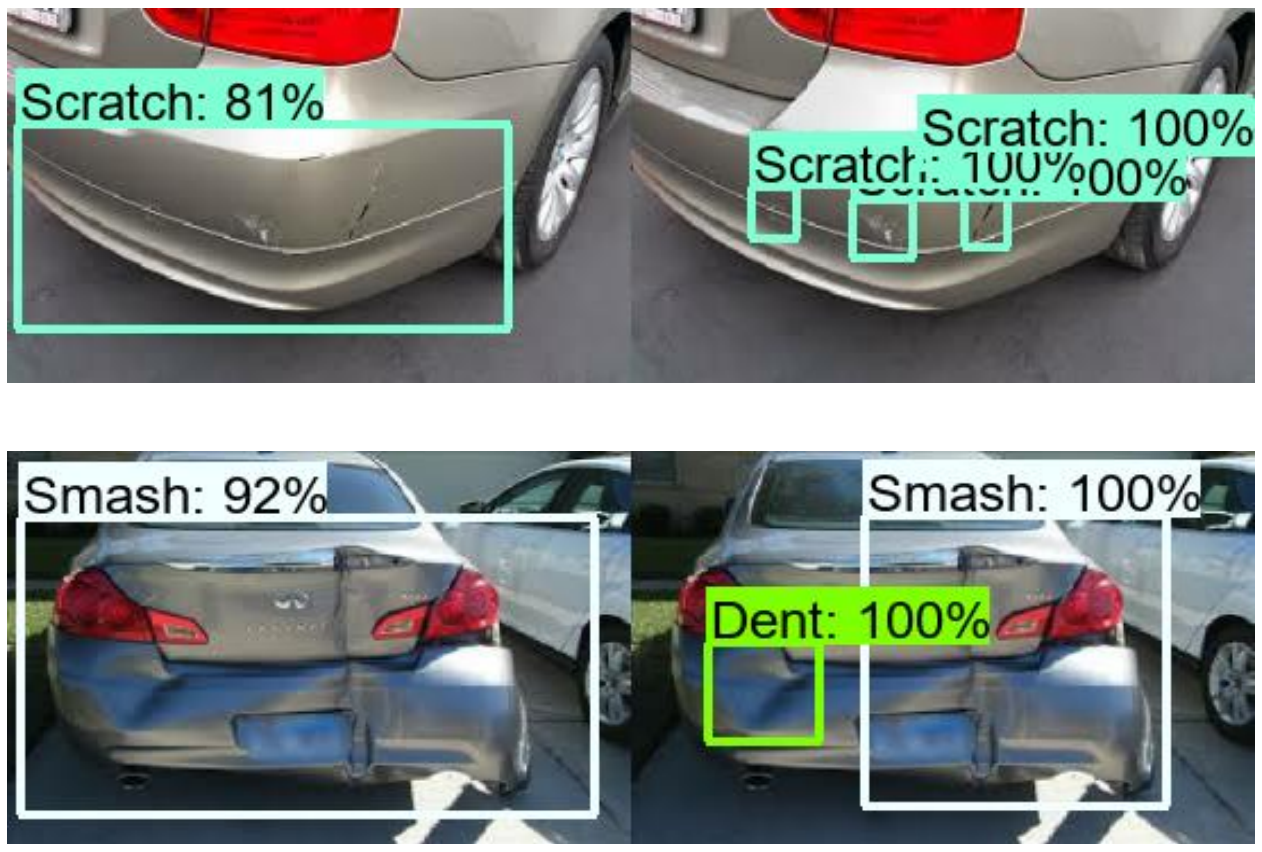

Fig. 6.5. Images on the left represent the predictions made by the model, while the right represents our annotations

The metric is assessed against the annotated test set we prepared, which in many instances separates classes, as can be seen in Fig 6.4, that the model seems to group into one main class. Therefore, when evaluating, these classifications are counted as errors even though the classification made by the model is valid. Therefore, this is the reason for its superior performance without a sufficient mAP score.

## VIII. CONCLUSIONS AND FUTURE SCOPE

In this research project we built a 2-layered AI model with the objective of optimizing performance, by filtering undamaged images and passing the filtered set to the object detection model for localization and classification into 4 categories, namely, scratches, dents, smashes, and glass shatters. The consolidated model evinces an optimized performance and thereby meets our purpose as per the defined use cases. The model architecture was put together with a variety of machine learning techniques to augment performance, like transfer learning, and data augmentation, that has built a foundation for the superior output of our model against a variety of images and videos.

Although our evaluation metrics, mAP of 0.5, do not back the significant performance of our model, the learning curve proves to be promising with an increase in training steps which we capped at 35K due to time constraints. Our classifier evinces an accuracy of 98%. On analyzing the results, the next steps for us would be to add more images that seemed to mislead the model to our dataset and perform a variety of hyperparameter tuning techniques. This project could be taken forward by integrating with two-staged models to segment the defects and determine damage levels that could precede the implementation of cost estimation. The label set can be expanded to accommodate a larger scope, while also projecting minute details like the position of the defect.

## ACKNOWLEDGMENTS

We would like to extend a vote of gratitude to our guides, and colleagues at Manipal Academy of Higher Education, Dubai Campus, recognizing the constant support and guidance they provided us with during the preparation of this paper. We would like to express our appreciation to the staff of the IT department for the valuable inputs and discussions that have supported this process.